%% file: SSHT.tex
\documentclass[sigconf]{acmart}
\usepackage{bbding}
\usepackage{eucal}
\usepackage{bbm}
\usepackage {subcaption}
\usepackage{booktabs}
\AtBeginDocument{%
  \providecommand\BibTeX{{%
    \normalfont B\kern-0.5em{\scshape i\kern-0.25em b}\kern-0.8em\TeX}}}

\begin{document}

\title{Learning Invariant Representation with Consistency and Diversity for Semi-supervised Source Hypothesis Transfer}

\author{Xiaodong Wang$^{1,}$*, Junbao Zhuo$^{2,\dag}$, Shuhao Cui$^2$, Shuhui Wang$^2$}
\affiliation{%
  \institution{\textsuperscript{\rm 1}School of Software and Microelectronics, Peking University}
  \institution{\textsuperscript{\rm 2}Key Lab of Intell. Info. Process., Inst. of Comput. Tech., CAS}
  }
  \email{wangxd220@gmail.com, junbao.zhuo@vipl.ict.ac.cn, {cuishuhao18s, wangshuhui}@ict.ac.cn}

  \makeatletter
\def\authornotetext#1{
  \if@ACM@anonymous\else
  \g@addto@macro\@authornotes{
    \stepcounter{footnote}\footnotetext{#1}}
  \fi}
\makeatother
\authornotetext{Xiaodong Wang was admitted as a PKU master student of 2021.}
\authornotetext{Junbao Zhuo is the corresponding author.}
\def\authors{Xiaodong Wang, Junbao Zhuo, Shuhao Cui, Shuhui Wang}
\renewcommand{\shortauthors}{Xiaodong Wang et al.}
\renewcommand{\shorttitle}{Semi-supervised Source Hypothesis Transfer}

\begin{abstract}
Semi-supervised domain adaptation (SSDA) aims to solve tasks in target domain by utilizing transferable information learned from the available source domain and a few labeled target data. However, source data is not always accessible in practical scenarios, which restricts the application of SSDA in real world circumstances. In this paper, we propose a novel task named Semi-supervised Source Hypothesis Transfer (SSHT), which performs domain adaptation based on source trained model, to generalize well in target domain with a few supervisions. In SSHT, we are facing two challenges: (1) The insufficient labeled target data may result in target features near the decision boundary, with the increased risk of mis-classification; (2) The data are usually imbalanced in source domain, so the model trained with these data is biased. The biased model is prone to categorize samples of minority categories into majority ones, resulting in low prediction diversity. To tackle the above issues, we propose Consistency and Diversity Learning (CDL), a simple but effective framework for SSHT by facilitating prediction consistency between two randomly augmented unlabeled data and maintaining the prediction diversity when adapting model to target domain. Encouraging consistency regularization brings difficulty to memorize the few labeled target data and thus enhances the generalization ability of the learned model. We further integrate Batch Nuclear-norm Maximization into our method to enhance the discriminability and diversity. Experimental results show that our method outperforms existing SSDA methods and unsupervised model adaptation methods on DomainNet, Office-Home and Office-31 datasets.
The code is available at https://github.com/Wang-xd1899/SSHT. 
\end{abstract}


\keywords{semi-supervised domain adaptation, unsupervised model adaptation, transfer learning.}


\maketitle

\input{intro}

\input{related}

\input{method}
\input{exp}
\input{conclusion}
\input{appendix}

\bibliographystyle{ACM-Reference-Format}
\bibliography{SSHT}

\end{document}

%% file: intro.tex
\section{Introduction}
\begin{figure}[h]
  \centering
  \includegraphics[width=0.8\linewidth]{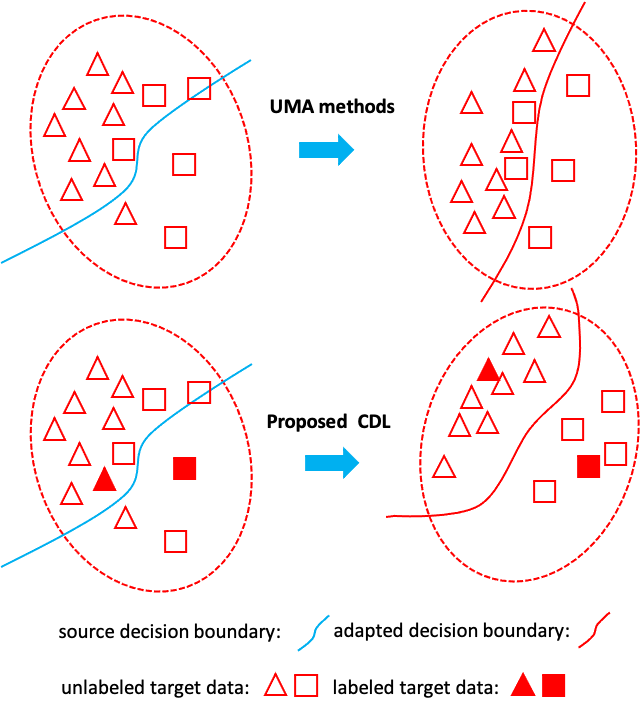}
  \caption{Comparison between unsupervised model adaptation methods (UMA methods, top) and our proposed CDL (SSHT method, bottom). UMA methods are prone to categorize the samples of minority categories into majority ones. Besides, the features of target domain samples lie near the decision boundary which may lead mis-classification. Our proposed CDL would push features of labeled samples far away from decision boundary and generates the decision boundary with large margin. Besides, our method could also maintain the prediction discriminability and diversity, improving its generalization ability.}
  \label{challenge}
\end{figure}
Deep learning methods have made a significant breakthrough with appreciable performance in a wide variety of applications under i.i.d. assumption. However, when training data and test data are not drawn from the same distribution, the trained model can not generalize well in test data. To deal with this domain shift problem, researchers resort to Unsupervised Domain Adaptation (UDA)~\cite{saenko2010adapting,pan2010domain,gong2012geodesic,tzeng2014deep,ganin2015unsupervised,long2015learning,long2017conditional,zhang2019bridging}. However, recent works~\cite{zhao2019learning,wu2019domain,combes2020domain} have shown that UDA does not guarantee good generalizations on the target domain. 
\begin{table*}
  \caption{The major differences among SSHT and other related DA settings.}
  \label{tab:diff}
  \begin{tabular}{cccc}
    \toprule
     &Source data&Trained source model&Target labeled data\\
    \midrule
    Unsupervised domain adaptation & \Checkmark & \XSolidBrush &\XSolidBrush \\
    Semi-supervised domain adaptation& \Checkmark & \XSolidBrush & \Checkmark\\
    Unsupervised model adaptation & \XSolidBrush & \Checkmark & \XSolidBrush\\
    SSHT &\XSolidBrush & \Checkmark & \Checkmark\\
  \bottomrule
\end{tabular}
\end{table*}
Especially when the marginal label distributions are distinct across domains, UDA methods provably hurt target generalization~\cite{zhao2019learning}. Besides, in many real-world applications, it is often feasible to at least obtain a small amount of labeled data from the target domain. Therefore, Semi-Supervised Domain Adaptation (SSDA)~\cite{donahue2013semi,saito2019semi,kim2020attract}, where the large amount of labeled source data and a small amount of labeled data from the target domain are available, has been given increasing attention.

In addition to utilizing a few labeled target samples, the major progress of SSDA has been developing improved methods for aligning representations between source and target in order to improve generalization. These methods span distribution alignment, for example by maximum mean discrepancy (MMD)~\cite{tzeng2014deep,long2015learning,yan2017mind}, domain adversarial training~\cite{ganin2015unsupervised,long2017conditional,zhang2019bridging}, and cycle consistent image transformation~\cite{liu2016coupled,hoffman2018cycada}. However, as revealed in a recent study~\cite{saito2019semi}, some UDA methods, e.g. DANN~\cite{ganin2015unsupervised} and CDAN~\cite{long2017conditional}, show no improvement or yield worse results than SSDA methods when trained on a few labeled target samples and source samples. Therefore, recent works focus on better leveraging the labeled and unlabeled target domain via min-max entropy \cite{saito2019semi}, meta-learning \cite{li2020online} and joint learning invariant representations and risks \cite{li2020learning}.


Despite its promising performance, SSDA is not always applicable in real-world scenarios as the source data is not always accessible for protecting the privacy in the source domain~\cite{liang2020we}. For example, many companies only provide the learned models instead of their customer data due to data privacy and security issues. Besides, the source datasets like videos or high-resolution images may be so large that it is not practical or convenient to transfer or retain them to different platforms~\cite{li2020model}. To overcome the absence of source data, unsupervised model adaptation (UMA) is investigated in \cite{liang2020we,li2020model}. UMA is tougher than UDA and inherits the challenges of UDA that the generalization ability on target domain may be not improved. Besides, without source data, it is hard to reduce domain discrepancy that the features of target domain samples lie near the decision boundary which may lead mis-classification, as shown in Fig~\ref{challenge}. To tackle these issues, in this paper we focus on a more realistic setting of Semi-supervised Source Hypothesis Transfer (SSHT), which has not been explored. The major differences among SSHT and other related problems are summarized in Table~\ref{tab:diff}.

SSHT is a more challenging task compared with SSDA as the source data is not accessible. In SSDA, even though the source domain is discrepant from target domain, the source labels are accurate for maintaining the discriminability of adapted model. While in SSHT, the insufficient labeled target data may result in target features lying near the decision boundary and increasing the risk of mis-classification. Besides, the source data are usually imbalanced that the trained model is prone to categorize the samples of minority categories into majority ones, which exhibits small prediction diversity. Such biased model trained on source data may not be well improved when transferred to target domain with only a few labeled samples, leading to poor generalization on target domain.

To tackle the above issues, we provide Consistency and Diversity Learning (CDL), a simple but effective framework for SSHT by encouraging prediction consistency on the unlabeled target data and maintaining the prediction diversity when adapting model to target domain. With two random data augmentations on an unlabeled image, the consistency regularization is achieved via interpolation consistency~\cite{zhang2017mixup,verma2019interpolation} or prediction consistency~\cite{berthelot2019mixmatch,sohn2020fixmatch}. We prefer Fixmatch~\cite{sohn2020fixmatch}, a simple but effective semi-supervised learning method. Fixmatch applies strong data augmentation~\cite{cubuk2020randaugment} to produce a wider range of highly perturbed images. Then regarding the predictions of weakly augmented images as pseudo labels, the consistency is achieved by training the model to categorize the strongly augmented images into the pseudo labels. Such consistency regularization makes the model harder to memorize the few landmarks and therefore enhances the generalization ability of the learned model. 





To maintain the prediction diversity, we integrate Batch Nuclear-norm Maximization (BNM)~\cite{cui2020towards} into our framework. As revealed in ~\cite{cui2020towards}, for a classification output matrix of a randomly selected batch data, the prediction discriminability and diversity could be separately measured by the Frobenius norm and rank of the matrix. As the nuclear-norm is an upperbound of the Frobenius-norm and a convex approximation of the matrix rank, encouraging Batch Nuclear-norm Maximization improves both discriminability and diversity. We argue that maintaining diversity is necessary since Fixmatch degrades diversity as it adopts only the samples with confident predictions higher than a predefined threshold for computing consistency regularization. Though such thresholding mechanism is helpful to mitigate the impact of incorrect pseudo labels, it will worsen the prediction diversity since samples of majority categories may exhibit larger prediction confidence.



We conduct extensive experiments on DomainNet, OfficeHome and Office-31. The experimental results show that the proposed CDL significantly outperforms state-of-the-art UMA methods and achieves comparable results against state-of-the-art SSDA methods. Ablation studies are presented to verify the contribution of each key component in our framework.






%% file: related.tex
\section{Related Work}
\subsection{Unsupervised Domain Adaptation}
The most deep neural network based Unsupervised Domain Adaptation (UDA) methods have made a success without any target supervision, which can be mainly categorizes into cross-domain discrepancy minimization based methods~\cite{tzeng2014deep, long2015learning,long2017deep} and adversarial adaptation methods~\cite{ganin2015unsupervised,long2017conditional,zhang2019bridging}. The popular discrepancy measurement, Maximum Mean Discrepancy (MMD), is firstly applied to one Fully-Connected (FC) layer of AlexNet in DDC~\cite{tzeng2014deep}. Deep Adaptation Network (DAN) \cite{long2015learning} further minimizes the sum of MMDs defined on several FC layers and achieves a better domain alignment. For a better discriminability in target domain, JAN~\cite{long2017deep} aligns the marginal and conditional distribution jointly based on MMD. Researcher also propose other discprepancy measurements such as correlation distance~\cite{sun2016deep} and Central Moment Discrepancy (CMD)~\cite{zellinger2017central} for UDA.

Inspired by adversarial learning, \cite{ganin2015unsupervised,long2017conditional,zhang2019bridging} impose the Gradient Reverse Layer (GRL) to better align domain distributions. In Domain Adversarial Neural Network (DANN)~\cite{ganin2015unsupervised}, the authors introduce a domain discriminator and learn features that are indistinguishable to the domain discriminator. In CDAN~\cite{long2017conditional}, the authors propose a novel conditional domain discriminator conditioned on domain-specific feature representations and classifier predictions, and implement discrepancy reduction via adversarial learning. To bridge the gaps between the theory and algorithm for domain adaptation, \cite{zhang2019bridging} present Margin Disparity Discrepancy (MDD) with rigorous generalization bounds, tailored to the distribution comparison with the asymmetric margin loss, and to the minimax optimization for easier training.

Some UDA methods focus on some characteristics of specific layer deep neural network for domain adaptation. In \cite{li2018adaptive}, the authors assume that the neural network layer weights learn categorical information and the batch norm statistics learn transferable information, so they propose AdaBN to modulating all Batch Normalization statistics from the source to target domain. In AFN~\cite{xu2019larger}, the authors reveal that the feature norms of target domain are much smaller than source domain and propose to adaptively increase the feature norms, which results in significant transfer gains. However, the prediction diversity is not explored that the model tends to push the examples near to the decision boundary, resulting error prediction accumulation. Batch Nuclear norm Maximization (BNM) \cite{cui2020towards}, adopted in this paper, maintains both discriminability and diversity, leading a promising result in several transfer learning tasks such as semi-supervised learning, domain adaptation and open domain recognition. 
\subsection{Semi-supervised Domain Adaptation}
Semi-Supervised Domain Adaptation (SSDA)~\cite{ao2017fast,donahue2013semi,yao2015semi,saito2019semi, xu2019d, kim2020attract,li2020online,li2020learning} is an extension of UDA with a few labeled target labels which achieves much better performance. Exploiting the few target labels allows better domain alignment compared to purely unsupervised approaches.
In \cite{donahue2013semi}, the authors impose smoothness constrains on the classifier scores over the unlabeled target data and lead to a better adaptation in conventional learning method. In \cite{yao2015semi}, the authors aim to learn a subspace to manifest the underlying difference and commonness between source and target domains, which reduces data distribution mismatch. In \cite{ao2017fast}, the authors estimate the soft label of the given labeled target sample with the source model and interpolated with the hard label for target model supervision. 
Work \cite{xu2019d} uses stochastic neighborhood embedding (d-SNE) to transform features into a common latent space for few-shot supervised learning, and use metric learning to improve the feature discrimination on the target domain.
In \cite{saito2019semi}, the authors point out that the weight vector of each class is an estimated prototype, and the entropy on target samples represents the similarity between prototypes and target features. Based on this assumption, they firstly maximize the the entropy of unlabeled target samples to move the weight vectors towards target data, and then update the feature extractor by minimizing the entropy of unlabeled target samples, leading to higher discriminability. Recently, work \cite{kim2020attract} raises a novel perspective of intra-domain discrepancy and propose a framework that consists attraction, perturbation, and exploration schemes to address the discrepancy.

\subsection{Model Adaptation}
Domain adaptation usually requires the large-scale source data, which is not practical due to the risk of violation of privacy in source domain. Therefore, the Model Adaptation (MA) \cite{liang2020we,li2020model,yang2020unsupervised,kundu2020universal,liang2021distill} is proposed to handle the domain adaptation when source data is unavailable. 

In \cite{liang2020we}, the source data is only exploited to train source model. Then they fine-tune the pre-trained model to learn source-like target representation. The key assumption in \cite{liang2020we} is that pre-trained model consists of a feature encoding module and a hypothetical classifier module. By fixing the classifier module, the fine-tuned encoding module can produce the better representations of target data, as source hypothesis encodes the distribution information of unseen source data. 
In \cite{li2020model}, the authors propose collaborative class conditional generative adversarial net, in which the prediction model is to be improved through generated target-style data. The prediction model can provide more accurate guidance for the generator that the generator and the prediction model can collaborate with each other. 
Liang et al~\cite{liang2021domain} develop two types of non-parametric classifiers, with an auxiliary classifier for target data to improve the quality of pseudo label when guiding the self-training process. In \cite{liang2020source}, the authors propose an easy-to-hard labeling transfer strategy, to improve the accuracy of less-confident predictions in target domain. 
Yang et al~\cite{yang2020unsupervised} handle this problem by deploying an additional classifier to align target features with the corresponding class prototypes of source classifier.  \cite{kundu2020universal} proposes a framework which exploits the knowledge of class-separability and enhances robustness to out-of-distribution samples. 
In \cite{liang2021distill}, the model provided as a black-box model to prevent generation techniques from leaking the individual information. These UMA methods inherit the challenges of UDA that the generalization ability on target domain may be not improved. Therefore, we raise SSHT to improve the generalization ability on target domain with just a few labeled target data.

%% file: method.tex
\section{Method}
\subsection{Semi-supervised Source Hypothesis Transfer}
\begin{figure*}[h]
  \centering
  \includegraphics[width=0.9\linewidth]{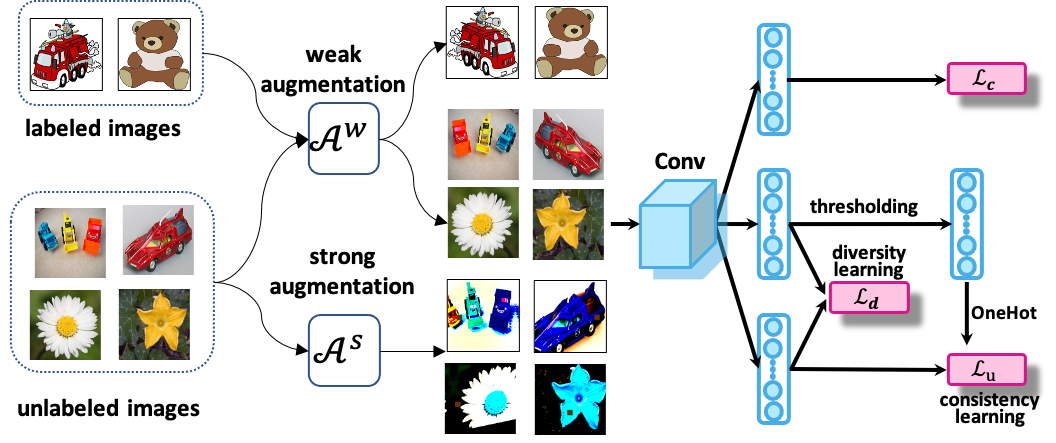}
  \caption{The proposed CDL framework for SSHT problem. Firstly, the unlabeled images are augmented with both weak and strong augmentations and fed to model. The prediction of weakly augmented images is use as supervision for the strongly augmented one to encourage the consistency of model prediction. We further encourage the prediction diversity by batch nuclear-norm maximization on outputs of all unlabeled augmented images.}
  \label{FR}
\end{figure*}
Common notations and definitions of Semi-supervised Source Hypothesis Transfer (SSHT) are introduced here. Suppose that there are labeled data $\mathcal{D}^{s} = \{(x_{i}^{s}, y_{i}^{s})\}_{i=1}^{N_{s}}$ in source domain. Similarly, we have unlabeled data $\mathcal{D}^{u} = \{u_i\}_{i=1}^{N_{u}}$ and a small set of labeled data $\mathcal{D}^{x} = \{(x_i, y_i)\}_{i=1}^{N_{x}}$ in target domain. $N_{u}$ is usually much larger than $N_{x}$, since the labeled data is more difficult to obtain.

Due to the data privacy, the source data in $\mathcal{D}^{s}$ is unavailable in SSHT. While we can leverage the model trained with source data. The model consists of a feature extractor and a classifier, where the parameters and weights are available. The goal of SSHT is to adapt the source model to target domain with only a few labeled target samples and unlabeled target samples. 
To address the issue of Semi-supervised Source Hypothesis Transfer, we provide a simple but effective framework that consists of the consistency learning (CL) and diversity learning (DL) modules. The overall framework is shown in Fig\ref{FR}. Firstly, the unlabeled images are augmented with both weak and strong augmentations. We feed the augmented data into the network and adopt the prediction results of weakly augmented images as supervision to train the strongly augmented ones for achieving prediction consistency. 
We maintain the prediction diversity by batch nuclear-norm maximization on outputs of all unlabeled augmented images. The source model is adapted in an end-to-end manner, and the collaboration between consistency learning and diversity learning enforces the decision boundary move away from labeled target samples towards unlabeled samples, improving the generalization ability of adapted model. 
\subsection{Consistency Learning}
The challenges of model adaptation is the absence of source data, which make it hard to estimate the distribution discrepancy between the two domains. Model adaptation without any labeled target sample is a complicated problem since the model would retain the decision boundary by the source information, and hard to be fine-tuned. With assistance of labeled target samples, the source model can learn some discriminative information in target domain. While the model may tends to overfit on labeled target data, resulting unreliable decision boundary.

To address the overfitting problem, some methods \cite{zhang2017mixup, verma2019interpolation, cubuk2020randaugment, sohn2020fixmatch} have been proposed based on data augmentation in a semi-supervised learning manner. 
Typical consistency regularization based methods \cite{sajjadi2016regularization, laine2016temporal} adopt the following loss:
\begin{displaymath}
  p(y\mid Augment_1(u); \theta) \approx  p(y\mid Augment_2(u); \theta)
\end{displaymath}
where $u$ is an unlabeled image. The $Augment_1$ and $Augment_2$ are different random augmentations. $\theta$ denotes parameters of model.

Besides, self training with pseudo-labeling is also a useful technique for semi-supervised learning. FixMatch~\cite{sohn2020fixmatch} is a combination of the two approaches to SSL: consistency regularization and pseudo-labeling. FixMatch utilizes a separate weak and strong augmentation when performing consistency regularization. Specifically, for each unlabeled sample $u \in \mathcal{D}^u$ in target domain, the weak augmentation $\mathcal{A}^w$ and strong augmentation $\mathcal{A}^s$ are defined as:
\begin{displaymath}
u_w = \mathcal{A}^w(u)\
\end{displaymath}
\begin{displaymath}
u_s = \mathcal{A}^s(u).
\end{displaymath}
The weak data augmentation $\mathcal{A}^w$ includes image flipping and image translation. And the strong data augmentation $\mathcal{A}^s$ utilizes the technique proposed in \cite{cubuk2020randaugment}. 
The consistency regularization incoporated with pseudo-labeling is implemented as treating the prediction of weakly augmented images as pseudo label and enforcing the prediction of strongle augmented ones towards the pseudo label. However, the pseudo labels may contain wrong labels, resulting in the error accumulation. Therefore, to mitigate the impact of incorrect pseudo labels, only samples with highly confident predictions are selected for consistency regularization. %
The consistency regularization loss on unlabeled images is defined as:
\begin{equation}
\mathcal{L}_{u} = \mathop{\mathbb{E}}_{u \sim \mathcal{D}^u}  \ \mathbbm{1}(max(p(u_w))>\tau)H(\hat{y}(u_w), p(u_s))
\end{equation}
where the $\tau$ is the threshold, and $\hat{y}(u_w)$ is the one-hot vector of $argmax(p(u_w))$. $H(p,q)$ denotes the cross-entropy between two distributions $p$ and $q$. 
By optimizing the consistency loss $\mathcal{L}_{u}$, the decision boundary will be pushed far from the labeled samples. Then the decision boundary enforces the model to be insensitive to the image perturbations and powerful in classifying unlabeled samples. 

To ensure the discriminability of model, we adopt the typical cross-entropy loss for the labeled target data $\mathcal{D}^x$. The classification loss $\mathcal{L}_{c}$ is defined as:
\begin{equation}
\mathcal{L}_{c} = \mathop{\mathbb{E}}_{(x, y) \sim \mathcal{D}^x} \  H(\hat{p}(x), y).
\end{equation}

The loss minimized by FixMatch is simply $\mathcal{L}_{c} + \lambda_u\mathcal{L}_{u}$ where $\lambda_u$ is a fixed scalar hyper-parameter denoting the relative weight of the unlabeled loss. 
\subsection{Diversity learning}
Though the selection mechanism is effective to mitigate the impact of incorrect pseudo labels, it will worsen the prediction diversity. Therefore we integrate an effective technique to maintain the discriminability and diversity of prediction. In domain adaptation, entropy minimization \cite{grandvalet2005semi} is widely adopted to enhance discriminability. However, simply minimizing entropy makes the trained model tend to classify samples near the decision boundary of the majority categories. Such unreliable classifiers will misclassify samples of minor categories which exhibits reduced prediction diversity. Though there are a few labeled target data in SSDA, it is insufficient to increase prediction diversity. 

To maintain the discriminability and diversity of prediction, we adopt Batch Nuclear-Norm Maximization (BNM)~\cite{cui2020towards}. Diversity could be measured by the number of response categories, which is the rank of the prediction matrix. And since the nuclear-norm is the convex approximation of the matrix rank, maximizing Batch Nuclear-norm will enlarge the rank, increasing the diversity. BNM is performed on the matrix of the classification responses for a batch unlabeled samples, without any supervision.

The loss function of BNM is defined as follow:
\begin{equation}
 \mathcal{L}_{bnm}=-\frac{1}{B}\vert \vert G(X)\vert \vert_\star
\end{equation}
where the $G(X)$ is the output matrix with respect to the input matrix $X$, and $B$ is the batch size of random samples. $\vert \vert \cdot \vert \vert_\star$ denotes the nuclear-norm, which is the sum of all the singular values in the matrix. In our settings, we have two augmented images, $\mathcal{A}^w$ and $\mathcal{A}^s$. Then the total loss for diversity learning is combined as follows:
\begin{equation}
	\begin{split}{
	\mathcal{L}_{d} = \mathop{\mathbb{E}}_{u_1,...,u_B \sim \mathcal{D}^u} -\frac{1}{B}(\vert \vert &G(\mathcal{A}^w([u_1,...,u_B]))\vert \vert_\star\\+ \vert \vert &G(\mathcal{A}^s([u_1,...,u_B]))\vert \vert_\star)
}\end{split}
\end{equation}
where $[.]$ denotes the calculation of concatenation. Minimizing the diversity loss can enforce the model to push the decision boundary into low density regions without losing diversity. In \cite{cui2020towards}, the authors reveal that the key insight of BNM may be sacrificing a certain level of the prediction hit-rate on majority categories, to enhance the prediction hit-rate on minority categories. Thus the diversity of prediction is retained. To maintain the discriminability, we minimize the diversity loss with the classification loss and consistency loss, and then model tends to produce more diverse and accurate prediction.
 \subsection{Training}
The total loss of the proposed CDL is defined as follows:                                                                                                                                                                 
 \begin{equation}
 \label{eq4}
\mathcal{L} = \mathcal{L}_{c} + \lambda_u\mathcal{L}_{u} + \lambda_d\mathcal{L}_{d},
\end{equation}
where the $\lambda_u$ and $\lambda_d$ control the trade-off between classification loss, consistency loss and diversity loss. The classification loss $\mathcal{L}_{c}$ provides accurate supervision for training model with high discriminability. The consistency regularization loss $\mathcal{L}_{u}$ prevents the model from overfitting on insufficient labeled target data, gaining better discriminability over unlabeled data. The diversity loss $\mathcal{L}_{d}$ could maintaining both the discriminability and diversity. The total loss encourages the trained model to generalize well on target domain.

%% file: exp.tex
\section{Experiment}
In this section, we conduct extensive experiments on typical domain adaptation benchmarks to verify the effectiveness of our method. For different tasks with the same source domain, we train a unique source model with the same source data. Then the source data are not used during adaptation. The results of recent state-of-the-art domain adaptation methods are presented for comparisons or as references since most of the methods are not applicable in the absence of source data during the adaptation process. 
\subsection{Datasets and settings}
\textbf{DomainNet} \cite{peng2019moment} is a recent benchmark dataset for large-scale domain adaptation with 345 classes across six domains. Following MME \cite{saito2019semi}, 7 scenario by selecting 4 domains (Real, Clipart, Painting, Sketch) and 126 classes are adopted here for fair comparison. The dataset is a new benchmark to evaluate semi-supervised domain adaptation methods.

\textbf{Office-Home} \cite{venkateswara2017deep} is a typical domain adaptation benchmark dataset, which consists of 15,500 images in 65 categories, mostly from an office or home environment. The images are sampled from four distinct areas including Art, Clipart, Product, and Real\_World with 65 classes. The methods are evaluated on 12 scenarios in total. 

\begin{table*}
	\caption{Accuracy of SSDA tasks on the DomainNet dataset ($\%$) (ResNet-34).}
	\label{tab:ssda_dn}
	\setlength{\tabcolsep}{3mm}{
		\begin{tabular}{c  c  c  c  c  c  c  c  c }
			\toprule
			Method &R  $\rightarrow$ C	& R  $\rightarrow$ P&P  $\rightarrow$ C&C  $\rightarrow$ S&S  $\rightarrow$ P&R  $\rightarrow$ S&P  $\rightarrow$ R&MEAN\\
			\midrule S+T \cite{he2016deep}&60.0 &	62.2 	&59.4 &	55.0 &	59.5 &	50.1 &	73.9 &	60.0 \\
			DANN \cite{ganin2015unsupervised}&	59.8 	&62.8 	&59.6 &	55.4& 	59.9 &	54.9 &72.2 &	60.7 \\
			ADR \cite{saito2017adversarial}	&60.7 &	61.9 &	60.7 	&54.4 &	59.9 &	51.1 	&74.2 &	60.4 \\
			CDAN \cite{long2017conditional}	&69.0 &	67.3 &	68.4& 	57.8 	&65.3 &	59.0& 	78.5 &	66.5 \\
			ENT \cite{grandvalet2005semi}&71.0&69.2	&71.1&60	&62.1&61.1&78.6&67.6\\ 
			MME \cite{saito2019semi}&72.2&69.7&71.7&61.8&66.8&61.9&78.5&68.9\\
			MixMatch \cite{berthelot2019mixmatch}	&72.6 	&68.8 	&68.7 	&62.7 	&67.1 	&65.5 	&78.7 	&69.2\\
			Meta-MME \cite{li2020online}&	73.5 	&70.3 &	72.8 	&62.8 &	68.0 &	63.8 &	79.2 &	70.1 \\
			BNM \cite{cui2020towards}	&72.7 &	70.2 &	72.5& 	63.9 &	68.8 	&63.0 	&80.3 &	70.2 \\
			GVBG \cite{cui2020gradually}	&73.3 	&68.7 	&72.9 	&65.3 	&66.6 	&68.5 	&79.2 	&70.6\\
			HDA \cite{cui2020hda}&73.9 &69.1 &73.0 &66.3 &67.5 &\textbf{69.5} &79.7 &71.3\\
			MME+ELP \cite{huang2020effective} &	74.9& 	72.1 	&74.4 &	64.3 	&69.7 &	64.9 	&81.0 &	71.6 \\
			APE \cite{kim2020attract}&76.6&72.1&\textbf{76.7}&63.1&66.1&67.8&79.4&71.7\\
			TML \cite{liu2020selective}&75.8&\textbf{74.5}&75.1&64.3&69.7&64.4&82.6&72.3\\
			ATDOC \cite{liang2021domain}	&\textbf{76.9} 	&72.5 	&74.2 	&66.7 	&70.8 	&64.6 	&81.2 	&72.4 \\
			\hline
			CDL (Ours) &75.5& 	73.0 &75.8 &	\textbf{67.2} 	&\textbf{71.5} &	65.8 &	\textbf{83.1} &	\textbf{73.1} \\
			\bottomrule
	\end{tabular}}
\end{table*}
\textbf{Office-31} \cite{saenko2010adapting} is a standard domain adaptation dataset which contains 4110 images from 31 categories with three domains: Amazon (A), with images collected from amazon.com, Webcam (W) and DSLR (D), with images shot by web camera and digital SLR camera respectively. Following TML~\cite{liu2020selective}, we evaluate the methods on two scenarios W $\rightarrow$ A and D $\rightarrow$ A for fair comparison.
\subsection{Implementation details}
\begin{table*}
	\caption{Accuracy of SSDA tasks on the Office-Home dataset ($\%$) (ResNet-34).}
	\label{tab:ssda_off}
	\begin{tabular}{c  c  c  c  c  c  c  c  c  c  c  c  c  c }
		\toprule
		Method&A $\rightarrow$ C&A  $\rightarrow$ P&A  $\rightarrow$ R&C  $\rightarrow$ A&C  $\rightarrow$ P&C  $\rightarrow$ R&P  $\rightarrow$ A&P  $\rightarrow$ C&P  $\rightarrow$ R&R  $\rightarrow$ A&R  $\rightarrow$ C&R  $\rightarrow$ P&MEAN\\
		\midrule S+T \cite{he2016deep}&	54.0 	&73.1 &	74.2 &	57.6 &	72.3 &	68.3 	&63.5 &	53.8 	&73.1 &	67.8 &	55.7 &	80.8 &	66.2 \\
		DANN \cite{ganin2015unsupervised}	&54.7 &	68.3 &	73.8 &	55.1 &	67.5 &	67.1 &	56.6 &	51.8 	&69.2 &	65.2 &	57.3 &	75.5 &	63.5 \\
		CDAN \cite{long2017conditional}	&59.2 &	74.1 	&74.1 	&60.5 	&74.5 &	70.7 &	61.4 	&58.1 &	76.8 &	67.1 &	61.4& 	80.7& 	68.2 \\
		BNM \cite{cui2020towards}&	62.2 	&78.6 &	78.9 &	65.0 &	78.0 &	77.8 	&65.2 &	60.4 &	80.3 	&69.0 &	63.4 	&84.2 &	71.9 \\
		ENT \cite{grandvalet2005semi}&61.3&79.5	&79.1&64.7&79.1&76.4&63.9&60.5&79.9&70.2	&62.6&85.7&71.9\\ 
		MME \cite{saito2019semi}&63.6&79&79.7&67.2&79.3&76.6&65.5&64.6&80.1&71.3&64.6&85.5	&73.0\\
		APE \cite{kim2020attract}&\textbf{63.9}&\textbf{81.1}&\textbf{80.2}	&66.6&79.9&76.8&66.1&\textbf{65.2}&\textbf{82}&73.4&\textbf{66.4}&86.2&74.0\\
		\hline
		CDL  (Ours)&63.0 &	80.2& 	80.1 &	\textbf{68.7} &	\textbf{82.0} 	&\textbf{78.8} 	&\textbf{68.5} &	62.7 &	81.7 	&\textbf{73.8} &	65.8 &	\textbf{86.5} 	& \textbf{74.3} \\
		\bottomrule
	\end{tabular}
\end{table*}
All the experiments are implemented with Pytorch \cite{paszke2017automatic}. For fair comparisons, we use the same backbones adopted in previous SSDA and UMA methods. For SSDA, ResNet-34 \cite{he2016deep} pre-trained on ImageNet \cite{deng2009imagenet} is widely adopted. Thus in the SSHT, we train the model based on pre-trained ResNet-34 in source domain to obtain the source model the same with UMA methods \cite{liang2020we, li2020model}. Following \cite{liu2020selective}, we use Vgg-16 \cite{simonyan2014very} pre-trained on ImageNet \cite{deng2009imagenet} on two scenarios W  $\rightarrow$ A and D  $\rightarrow$ A of Office-31 to evaluate methods.

All the SSDA and SSHT tasks are in the three-shot setting. For the UMA, we use the pre-trained ResNet-50~\cite{he2016deep} as the backbone, and then train the model on source domain. Following \cite{liang2020we, li2020model}, we split the labeled source data into a training set and a validation set, with the ratio of $9:1$. The provided model is trained on the training set, and be validated on validation set to avoid overfitting to the source data. The methods such like ENT \cite{grandvalet2005semi}, MME \cite{saito2019semi} and BNM \cite{cui2020towards} are implemented with the same hyper-parameters as \cite{cui2020towards}. We use the SGD optimizer with learning rate 0.005, nesterov momentum 0.9, and weight decay 0.0005. We set $\lambda_u$ to 2.5 and $\lambda_d$ to 1.0 for all datasets. We set batch size to 48, 96, 48 in Office-Home, DomainNet and Office-31, respectively. We train the proposed CDL with 30 epochs in total. The threshold $\tau$ is set to 0.8 for selecting samples with highly confident predictions. More details can be seen in our released codes. 
\subsection{Compared methods}

\begin{table}
	\caption{Accuracy of two SSDA tasks on the Office-31 dataset ($\%$) (Vgg-16). }
	\label{tab:vgg}
		\begin{tabular}{c  c  c  c  c  c  c  c  }
			\toprule METHOD &W  $\rightarrow$ A	& D  $\rightarrow$ A&MEAN\\
			\midrule CDAN \cite{long2017conditional}&	74.4 &	71.4 &	72.9 \\
S+T	\cite{he2016deep}&73.2 &	73.3 &	73.3 \\
ADR \cite{saito2017adversarial}&	73.3& 	74.1 	&73.7 \\
DANN \cite{ganin2015unsupervised}&	75.4& 	74.6 &	75.0\\ 
ENT \cite{grandvalet2005semi}&	75.4 	&75.1 &	75.3 \\
MME	 \cite{saito2019semi}&76.3 &	77.6 &	77.0 \\
TML \cite{liu2020selective}&	76.6& 	77.6 	&77.1 \\
\midrule CDL (Ours)&	\textbf{78.0} &	\textbf{78.1}& 	\textbf{78.1} \\
\bottomrule
	\end{tabular}
\end{table}
\begin{table*}
  \caption{Accuracy of UMA tasks on the Office-Home dataset ($\%$) (ResNet-50). }
  \label{tab:uda_off}
  \resizebox{\linewidth}{!}{
  \begin{tabular}{c  c  c  c  c  c  c  c  c  c  c  c  c  c }
    \toprule
     Method&A $\rightarrow$ C&A  $\rightarrow$ P&A  $\rightarrow$ R&C  $\rightarrow$ A&C  $\rightarrow$ P&C  $\rightarrow$ R&P  $\rightarrow$ A&P  $\rightarrow$ C&P  $\rightarrow$ R&R  $\rightarrow$ A&R  $\rightarrow$ C&R  $\rightarrow$ P&MEAN\\
     \midrule S \cite{he2016deep} &44.6 &67.3 &74.8 &52.7 &62.7 &64.8 &53.0 &40.6 &73.2 &65.3 &45.4 &78.0 &60.2 \\

    DANN \cite{ganin2015unsupervised}	&45.6 &	59.3 &	70.1 &	47.0 &	58.5 	&60.9 &	46.1& 	43.7 &	68.5 &	63.2& 	51.8 &	76.8 &	57.6 \\
DAN \cite{long2015learning}	&43.6 &	57.0 	&67.9 &	45.8 &	56.5 	&60.4 &	44.0 &	43.6 	&67.7 &	63.1 &	51.5 	&74.3 &	56.3 \\
CDAN \cite{long2017conditional}&	50.7 &	70.6 	&76.0 &	57.6 	&70.0 &	70.0 	&57.4 &	50.9 &	77.3 	&70.9 &	56.7 &	81.6 &	65.8 \\
SAFN \cite{xu2019larger}&	52.0& 	71.7& 	76.3 &	64.2& 	69.9& 	71.9 &	63.7 	&51.4 &	77.1 &	70.9& 	57.1 &	81.5& 	67.3\\ 
MDD	\cite{zhang2019bridging} &54.9 &	73.7 	&77.8 &	60.0 	&71.4 &	71.8 &	61.2 	&53.6 &	78.1 &	72.5 &	60.2 &	82.3 	&68.1 \\
SHOT \cite{liang2020we} &	57.3& 	78.5 	&81.4 &	67.9 	&78.5 &	78.0 &	68.1 &	56.1& 	82.1 	&73.4 &	59.6 &	84.4 &	72.1 \\
ATDOC \cite{liang2021domain}&	58.3 	&78.8 &	82.3 &	69.4 &	78.2 &	78.2 	&67.1 &	56.0 	&82.7 &	72.0 &	58.2 	&85.5 &	72.2 \\
SHOT++ \cite{liang2020source}	&58.1 &	79.5 &	\textbf{82.4} &	68.6 	&79.9 &	79.3 	&68.6 &	57.2 	&83.0 &	74.3& 	60.4& 	85.1 &	73.0 \\
\hline
CDL (Ours)&	\textbf{61.2}& 	\textbf{83.8}& 	81.9 	&\textbf{73.3} &	\textbf{85.0} 	&\textbf{81.3} &	\textbf{71.0} &	\textbf{61.9} &	\textbf{83.2} 	&\textbf{76.2} &	\textbf{63.0} &	\textbf{87.0} 	&\textbf{75.7} \\
  \bottomrule
\end{tabular}}
\end{table*}
\begin{figure}[h]
  \centering
  \begin{subfigure}[b]{0.235\textwidth}
        \includegraphics[width=\textwidth]{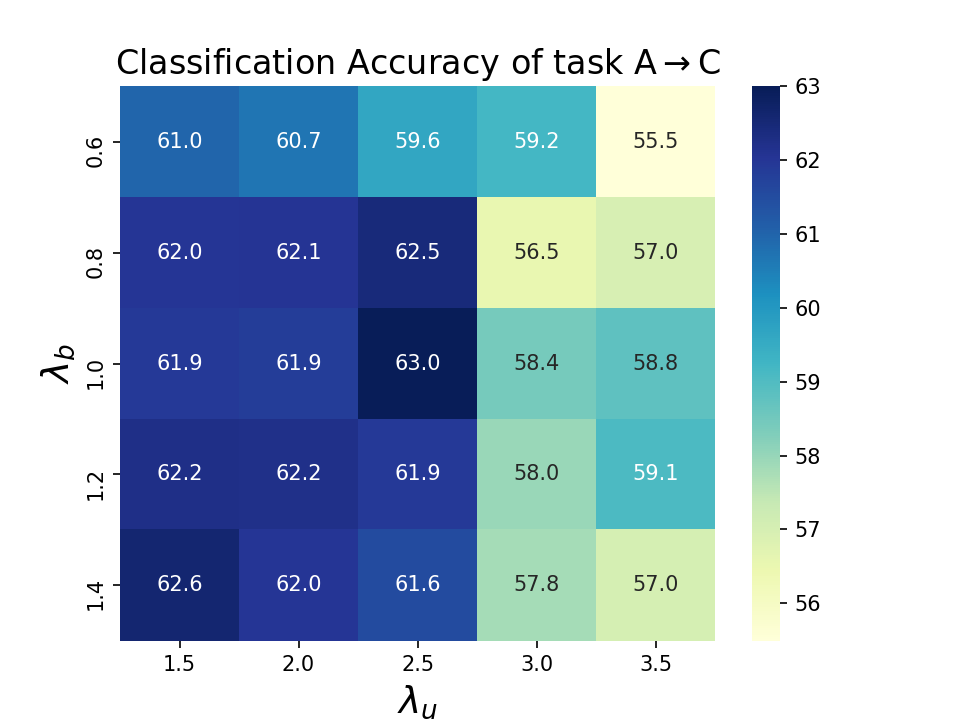}
        \caption{A $\rightarrow$ C}
        \label{fig:ps_ac}
    \end{subfigure}
    \begin{subfigure}[b]{0.235\textwidth}
        \includegraphics[width=\textwidth]{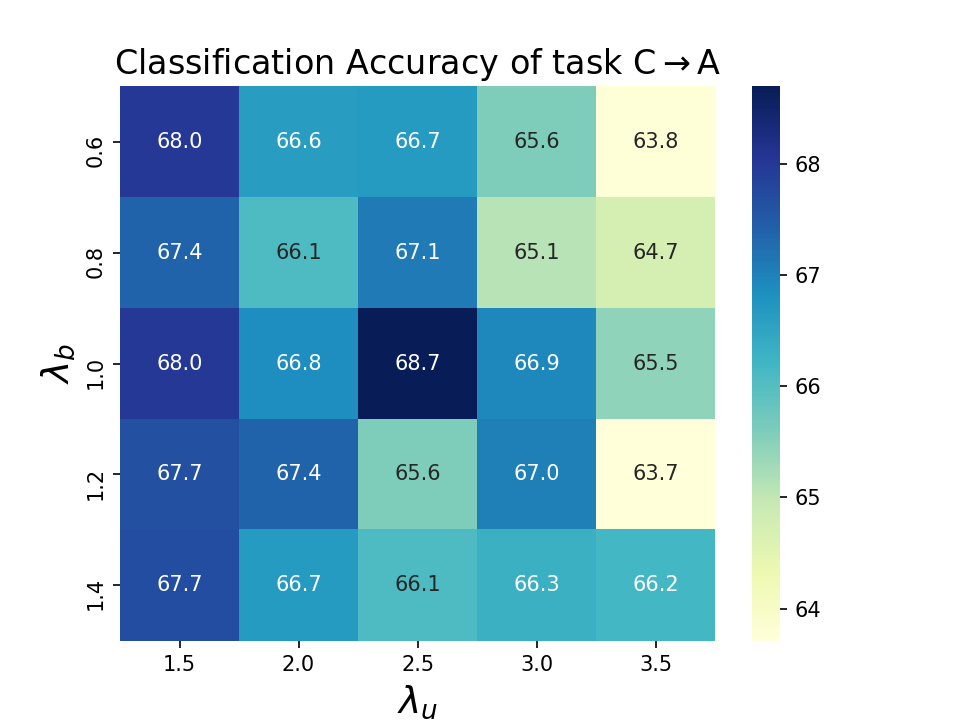}
        \caption{C $\rightarrow$ A}
        \label{fig:ps_ca}
    \end{subfigure}
 \caption{Sensitivity analysis of hyper-parameters $\lambda_u$ and $\lambda_b$ in SSHT tasks on Office-Home (ResNet-34). }
\label{fig:ps}
\end{figure}

\begin{table}
	\caption{Accuracy of data-based and model-based setting on the Office-Home dataset ($\%$) (ResNet-34). }
	\label{tab:model}
	\resizebox{\linewidth}{!}{
		\begin{tabular}{c  c  c  c  c  c  c  c  }
			\toprule
			METHOD &A  $\rightarrow$ C	& A  $\rightarrow$ P&A  $\rightarrow$ R&C  $\rightarrow$ A&C  $\rightarrow$ P&C  $\rightarrow$ R&MEAN\\
			\midrule SSL+CL &27.9 &64.5 &	51.0 &37.8 &62.4 &48.2 &48.6\\
			\midrule ENT (w/ data)\cite{grandvalet2005semi}&61.3	&79.5&79.1&64.7&79.1&76.4&73.4\\ 
			ENT (w/ model)\cite{grandvalet2005semi}	&58.3&78.0&78.5	&63.4&77.4&75.1&71.8\\
			\midrule
			MME (w/ data)\cite{saito2019semi}&63.6	&79.0&79.7&67.2&79.3&76.6&74.2\\
			MME (w/ model)\cite{saito2019semi}&51.4&69.5&67.4&54.7&68.5&63.6&62.5\\
			\midrule
			BNM (w/ data)\cite{cui2020towards}&62.2	&78.6&78.9&65.0&78.1&77.8&73.4\\
			BNM (w/ model) \cite{cui2020towards}&61.0&78.8&\textbf{80.2}&65.6&78.9&78.0&73.8\\
			\hline
			CDL (w/ data)&\textbf{63.0} &\textbf{81.0}&80.1& 67.2 &80.6 &\textbf{80.0} &75.3 \\
			\textbf{CDL} (w/ model)&\textbf{63.0} &	80.2 &	80.1 &	\textbf{68.7} &	\textbf{82.0}& 78.8 	&\textbf{75.4} \\
			\bottomrule
	\end{tabular}}
\end{table}

\begin{figure}[h]
	\centering
	\begin{subfigure}[b]{0.235\textwidth}
		\includegraphics[width=\textwidth]{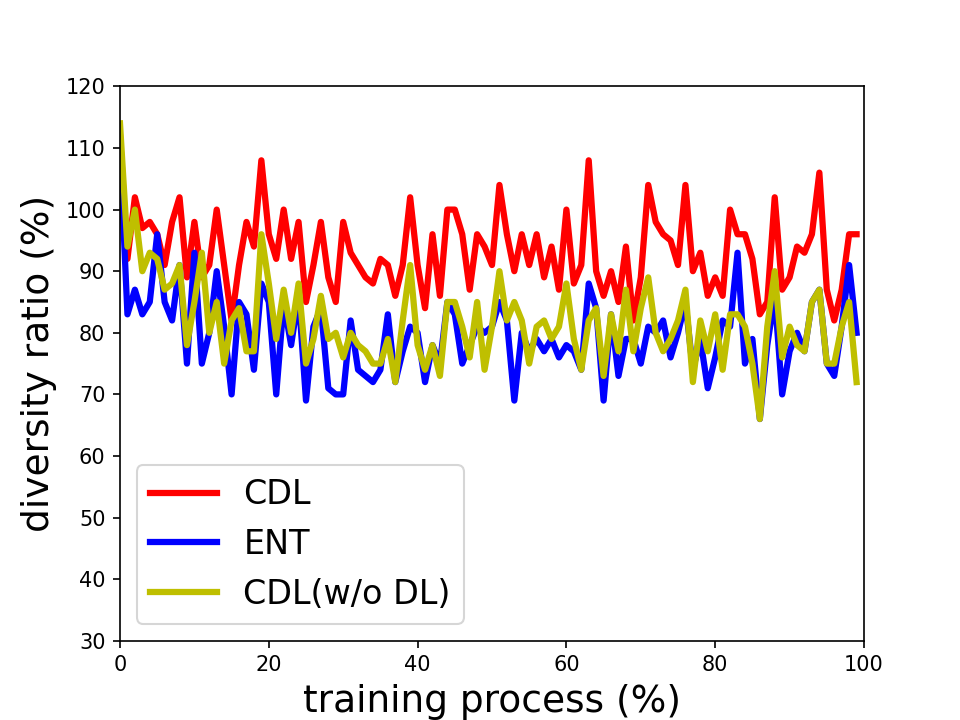}
		\caption{A $\rightarrow$ C}
		\label{fig:dv_ac}
	\end{subfigure}
	\begin{subfigure}[b]{0.235\textwidth}
		\includegraphics[width=\textwidth]{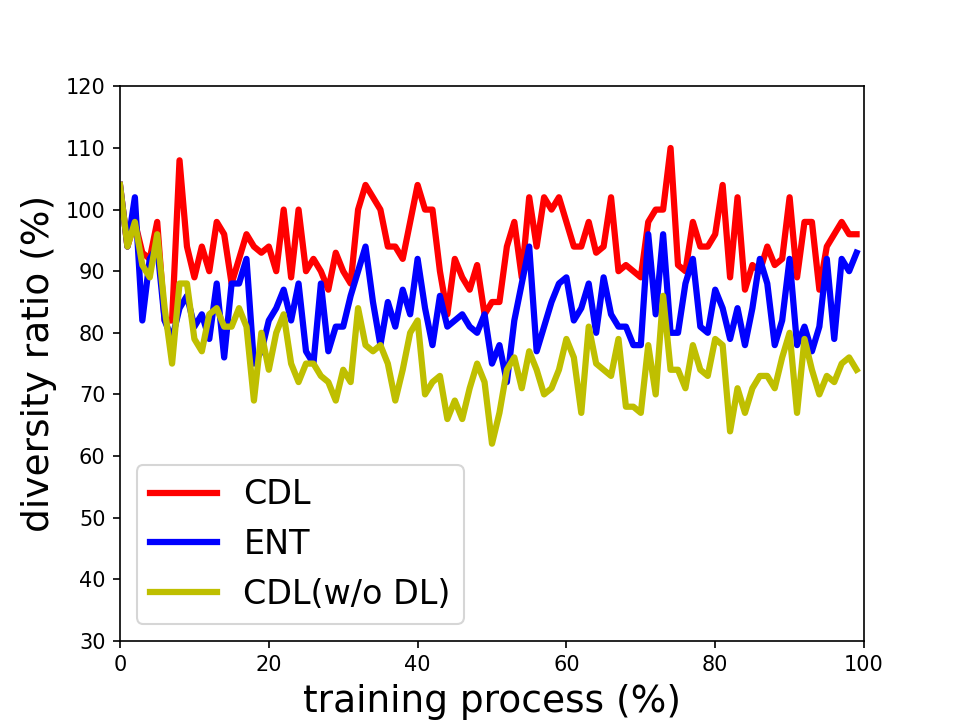}
		\caption{P $\rightarrow$ A}
		\label{fig:dv_pa}
	\end{subfigure}
	\caption{Diversity ratio on Office-Home for domain adaptation, calculated as the predicted diversity divided by the ground truth diversity. The predicted (ground truth) diversity is measured by the number of predicted (ground truth) categories in randomly sampled batches}
	\label{fig:dv}
\end{figure}

\begin{table*}
  \caption{Ablation study on the Office-Home dataset ($\%$) (ResNet-34). }
  \label{tab:ablation}
  \resizebox{\linewidth}{!}{
  \begin{tabular}{c  c  c  c  c  c  c  c  c  c  c  c  c  c }
    \toprule
     Method&A $\rightarrow$ C&A  $\rightarrow$ P&A  $\rightarrow$ R&C  $\rightarrow$ A&C  $\rightarrow$ P&C  $\rightarrow$ R&P  $\rightarrow$ A&P  $\rightarrow$ C&P  $\rightarrow$ R&R  $\rightarrow$ A&R  $\rightarrow$ C&R  $\rightarrow$ P&MEAN\\
    \midrule CDL (w/o CL)	&60.9 &	79.0 &	\textbf{80.3} &	66.1 &	79.0 	&\textbf{78.9} &	66.6 	&61.7 &	80.4 &	70.3 &	64.3& 	85.2 &	72.7 \\
    
CDL (w/o DL) &	54.9 	&77.4 &	75.5 &	62.4 	&76.8 	&73.6 &	62.3 	&57.5 &	76.4 &	68.1 &	59.0 	&83.3 &	68.9 \\
    
CDL (Ours)&\textbf{63.0} &	\textbf{80.2} &	80.1 &	\textbf{68.7} &	\textbf{82.0} &	78.8& 	\textbf{68.5} &	\textbf{62.7}& 	\textbf{81.7} &	\textbf{73.8} 	&\textbf{65.8} &	\textbf{86.5} &	\textbf{74.3} \\
 \bottomrule
    \end{tabular}}
\end{table*}
\textbf{SSDA.} We compare our method with SSDA methods and some UDA methods compared in previous works\cite{saito2019semi, kim2020attract}. DANN \cite{ganin2015unsupervised} is a popular method employing a domain classifier to match feature distribution. ADR \cite{saito2017adversarial} utilizes adversarial dropout regularization to encourage the generator to output more discriminative features for the target domain. CDAN \cite{long2017conditional} performs distribution alignment by a class-conditioned domain discriminator. All the above methods are implemented and evaluated under the SSDA setting. S+T \cite{he2016deep} is a vanilla model trained with the labeled source and labeled target data without using unlabeled target data. BNM \cite{cui2020towards} is a method using nuclear-norm maximization in each batch samples for maintaining discriminability and diversity of prediction. ENT \cite{grandvalet2005semi} could be applied to SSDA by the entropy minimization. MME \cite{saito2019semi} adopts a minimax game on the entropy of unlabeled data. APE\cite{kim2020attract} aligns features via alleviation of the intra-domain discrepancy. MixMatch \cite{berthelot2019mixmatch} is a method to deal with semi-supervised-learning, and can also be applied on SSDA. Meta-MME \cite{li2020online} incorporates meta-learning to search for better initial condition in domain adaptation. MME+ELP \cite{huang2020effective} tackles the problem of lacking discriminability by using effective inter-domain and intra-domain semantic information propagation. GVBG \cite{cui2020gradually} proposes a novel gradually vanishing bridge to connect either source or target domain to intermediate domain. HDA \cite{cui2020hda} devises a heuristic framework to conduct domain adaptation.
TML \cite{liu2020selective} proposes a novel reinforcement learning based selective pseudo-labeling method to deal with SSDA, which employes deep Q-learning to train an agent to select more representative and accurate pseudo-labeled samples for model training. ATDOC \cite{liang2021domain} develops two types of non-parametric classifiers, with an auxiliary classifier for target data to improve the quality of pseudo label. For fair comparison, all the methods have the same backbone architecture with our method.

\textbf{Unsupervised model adaptation.} Except for DANN \cite{ganin2015unsupervised}, ATDOC \cite{liang2021domain}, and CDAN \cite{long2017conditional}, we compare our method with other UDA methods such as DAN \cite{long2015learning}, MDD \cite{zhang2019bridging}, SAFN \cite{xu2019larger}, SHOT \cite{liang2020we}, and SHOT++ \cite{liang2020source}. DAN \cite{long2015learning} utilizes a multi-kernel selection method for better mean embedding matching and adapts in multiple layers to learn more transferable features. MDD~\cite{zhang2019bridging} is a measurement with rigorous generalization bounds, tailored to the distribution comparison with the asymmetric margin loss, and to the minimax optimization for easier training. SAFN \cite{xu2019larger} proposes a norm adaptation to well discriminate the source and target features. SHOT \cite{liang2020we} addresses unsupervised model adaptation with self-supervision learning. And the SHOT++ \cite{liang2020source} proposes a labeling transfer strategy to improve the accuracy of less-confident predictions on the basis of SHOT. 
\subsection{Results}

\textbf{Comparison with SSDA methods}. The results of our CDL in the SSHT setting is compared with other methods which could access the source data. The comparison results on DomainNet and Office-Home are shown in Table \ref{tab:ssda_dn} and Table \ref{tab:ssda_off}, respectively. As for DomainNet, our CDL outperforms state-of-the-art method ATDOC \cite{liang2021domain} by 0.7\% in average. In the task P $\rightarrow$ R, our CDL significantly outperforms the ATDOC by 1.9\%. Specifically, CDL outperforms ATDOC in 6 transfer tasks over 7 tasks. In general comparison with others, our method achieve the best results in 3 tasks. Although our method shows weakness in some tasks such like R $\rightarrow$ S, it outperforms other methods in average. As shown in Table \ref{tab:ssda_off}, we can observe that our method CDL achieves comparable results against state-of-the-art SSDA methods on Office-Home, moreover, shows the best accuracy in 6 tasks over 12 tasks. We also evaluate our method in Office-31 for the setting in \cite{liu2020selective}. The comparison results on Office-31 in Table \ref{tab:vgg} shows that our method CDL based on model outperforms significantly the other methods based on source data in both two scenarios, and it outperforms the state-of-the-art TML by 1.0\% in average.
It is worthy of noting that the accurate labeled source data are accessible for SSDA methods, making it easier to transfer compared with SSHT. Despite the absence of source data, the superiority of CDL over state-of-the-art SSDA methods validates the effectiveness of CDL. 

\textbf{Comparison with UMA methods}. The difference between the SSHT and UMA is that SSHT has extra labeled data for model adaptation. We compare our CDL on Office-Home with previous methods tailored or applicable for UMA. The results in Table \ref{tab:uda_off} show that our CDL outperforms state-of-the-art method SHOT++ by 2.7\% in average. Our CDL yields the great improvement by effectively learning invariant representation with a few target supervisions. It is worthy of noting that CDL outperforms SHOT++ in 11 transfer tasks over the total 12 tasks. The superiority of CDL over UMA methods validates that even with few labeled target data, the performance can be significantly improved. 

\textbf{Effectiveness of adaptation}. To validate that our method is effective to the SSHT problem, we evaluate our method on six closed-set SSDA tasks without source data.
The results are shown in Table \ref{tab:model}. SSL+CL denotes semi-supervised learning with consistency learning. ENT (w/ data) denotes adaption from source data. ENT (w/ model) stands for adaptation from source model. Our CDL is the framework aims to address the SSHT problem, and can also be applied on SSDA problem. Compare the SSL+CL with others, it proves that the adaptation based on data or model is superior than only semi-supervised learning with consistency learning in target domain. It is shown that BNM can handle the difference between data and model while accuracies of others are decreased. And our CDL shows superiority than others in both SSDA and SSHT. 

\subsection{Ablation Study}
Since our CDL framework comprises a simple combination of consistency learning (CL) and diversity learning (DL), we perform an extensive ablation study to better understand why it is able to perform favorably against state-of-the-art methods in SSDA and UMA. We evaluate two variants of CDL: (1) \textbf{CDL (w/o CL)}, which denotes that we adapt the model without learning the consistency of unlabeled images, only by optimizing the classification loss of labeled images and the loss of diversity. (2) \textbf{CDL (w/o DL)}, which only optimizes the loss of consistency learning and classification loss of labeled images in the training process. The results of ablation study are shown in Table \ref{tab:ablation}. We can observe that the two components are designed reasonably and when any one of the two components is removed, the performance degrades. It is noteworthy that the CDL (w/o CL) outperforms the full CDL method on two tasks, showing the effectiveness of maintaining diversity in model adaptation. Our CDL combines the CL and DL and obtains 1.6\% improvement in average, which validates the effectiveness of CDL. 

\vspace{-1ex}
\subsection{Further remarks}
\textbf{Effectiveness of maintaining diversity}. To validate that our method could maintain the diversity in model adaptation, we compared our method with the our variant CDL (w/o DL) and entropy minimization. We show the diversity ratio in Office-Home on tasks of A $\rightarrow$ C and P $\rightarrow$ A in Fig \ref{fig:dv}. The diversity is measured by the number of predicted categories in randomly sampled batch. Thus the diversity ratio is calculated as the predicted diversity divided by the ground truth diversity. As shown in Fig \ref{fig:dv_ac}, the diversity ratio of CDL is larger than others, and the CDL (w/o DL) shows the comparable diversity loss in task A $\rightarrow$ C. As shown in Fig \ref{fig:dv_pa}, the CDL (w/o DL) shows low diversity ratio, while our CDL still maintain the large diversity ratio in the harder task P $\rightarrow$ A. 

\begin{table*}
    \caption{Accuracy ($\%$) Oon VisDA-2017 for UDA (ResNet-34), (MME and CDL are tested in SSHT). }
    \label{tab:visda}
    \resizebox{\linewidth}{!}{
    \begin{tabular}{c  c  c  c  c  c  c  c  c  c  c  c  c  c }
      \toprule
       Method&plane&bcybl&bus&car&horse&knife&mcyle&persn&plant&sktb&train&trunk&mean\\
      \midrule ResNet&55.1 &	53.3 &61.9 &	59.1 &	80.6 	&17.9 &	79.7 	&31,2 &	81.0 &	26.5 &	73.5& 	8.5 &	52.4 \\
      
  CDAN &	85.2 	&66.9 &	83.0 &	50.8 	&84.2 	&74.9 &	88.1 	&74.5 &	83.4 &	76.0 &	81.9 	&38.0 &	73.9 \\
      
  AFN&93.6 &61.3 &	84.1 &	70.6 &	94.1&	79.0& 91.8 &	79.6& 89.9 &	55.6 	&89.0&24.4 &	76.1 \\
  \midrule MME&  94.1&	83.0&	79.3 &	52.5 	&87.4 &	96.4 &	77.2 &	79.2 &	88.4 	&78.2 &	87.6 &	45.6 &	79.1  \\
  CDL &\textbf{97.5} &	\textbf{88.9} &	\textbf{84.9} &	\textbf{84.2} &	\textbf{96.6} &	\textbf{97.5} &	\textbf{92.3} &	\textbf{84.3} &	\textbf{96.5} &	\textbf{95.5} &	\textbf{92.1} 	&\textbf{51.5} &	\textbf{88.5} \\

   \bottomrule
      \end{tabular}}
  \end{table*}

\textbf{Parameter sensitivity}. We evaluate the effects of the parameters $\lambda_u$ and $\lambda_d$ in SSHT task, which control the trade off between consistency loss, diversity loss and classification loss. We evaluate several combination of $\lambda_u$ and $\lambda_d$ in two tasks A $\rightarrow$ C and C $\rightarrow$ A on Office-Home. As shown in Figure \ref{fig:ps}, we see that appropriate combination of $\lambda_u$ and $\lambda_d$ results in good transfer performance in model adaptation. This justifies our motivation of learning invariant representation with encouraging consistency and maintaining diversity by the proposed method, as a good trade-off among them can promote transfer performance.

%% file: conclusion.tex
\section{Conclusion}
In this paper, we propose a novel Semi-supervised Source Hypothesis Transfer (SSHT) task to fully utilize a few labeled target data and inherit knowledge of source model. The insufficient labeled target data may increase the risk of mis-classification in target domain and reduce the prediction diversity. To tackle these issues, we present Consistency and Diversity Learning (CDL) framework for SSHT. By encouraging consistency regularization between two random augmentations of unlabeled data, the model can generalize well in target domain. In addition, we further integrate Batch Nuclear-norm Maximization (BNM) to enhance the diversity. Experimental results on multiple domain adaptation benchmarks show that our method outperforms existing state of the art SSDA methods and unsupervised model adaptation methods. 

%% file: appendix.tex
\section{Appendix}
  We conduct the experiment on Visda-2017 for UDA. To adopt the SSHT setting, 
  we use three labeled samples in validation domain of Visda, and adpat the model 
  to validation domain. The result is shown
   in Table \ref{tab:visda}. Our \textbf{CDL} archives a better average accuracy 
   among the above methods.